\documentclass[10pt,twocolumn,letterpaper]{article}

\usepackage{cvpr}
\usepackage{times}
\usepackage{epsfig}
\usepackage{graphicx}
\usepackage{amsmath}
\usepackage{amssymb}
\usepackage{subfigure}
\usepackage{multirow}

\usepackage[breaklinks=true,bookmarks=false]{hyperref}

\cvprfinalcopy 

\def\cvprPaperID{****} 

\begin{document}

\title{Best Vision Technologies Submission to ActivityNet Challenge 2018\\---Task: Dense-Captioning Events in Videos}

\author{Yuan Liu and Moyini Yao\\
         Best Vision Technologies Co., Ltd, Beijing, China\\
                                {\tt\small \{yuan.liu, ting.yao\}@y-cv.com}
}

\maketitle

\begin{abstract}
This note describes the details of our solution to the dense-captioning events in videos task of ActivityNet Challenge 2018. Specifically, we solve this problem with a two-stage way, i.e., first temporal event proposal and then sentence generation. For temporal event proposal, we directly leverage the three-stage workflow in \cite{yao2017msr,zhao2017temporal}. For sentence generation, we capitalize on LSTM-based captioning framework with temporal attention mechanism (dubbed as LSTM-T). Moreover, the input visual sequence to the LSTM-based video captioning model is comprised of RGB and optical flow images. At inference, we adopt a late fusion scheme to fuse the two LSTM-based captioning models for sentence generation.
\end{abstract}

\section{Sentence Generation Model}
\begin{figure}[!tb]
\centering {\includegraphics[width=0.5\textwidth]{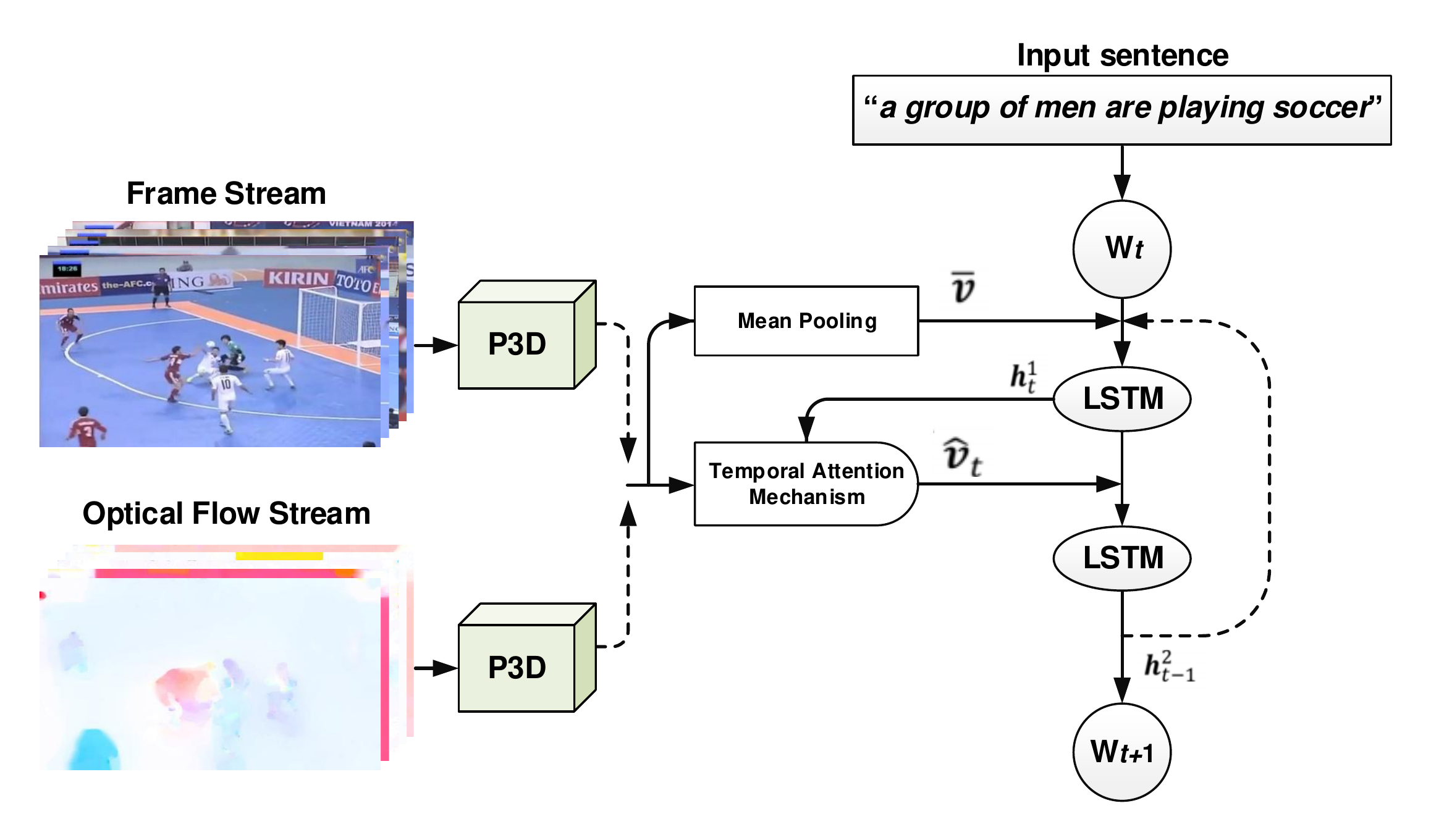}}
\caption{The sentence generation model in our system for dense-captioning events in videos task.}
\label{fig:figall}
\end{figure}

Inspired from the recent successes of LSTM based sequence models leveraged in image/video captioning \cite{donahue2015long,li2018jointly,pan2016jointly,pan2017seeing,pan2017video,venugopalan2015sequence,Xu:ICML15,yao2015describing,yao2017novel,yao2017boosting}, we formulate our sentence generation model in an end-to-end fashion based on LSTM which encodes the input frame/optical flow sequence into a fixed dimensional vector via temporal attention mechanism and then decodes it to each target output word. An overview of our sentence generation model is illustrated in Figure \ref{fig:figall}.

\begin{table*}[!tb]
\centering
\caption{Performance on ActivityNet captions validation set, where B@$N$, M, R and C are short for BLEU@$N$, METEOR, ROUGE-L and CIDEr-D scores. All values are reported as percentage (\%).}
\label{table:tabdense}
\begin{tabular}{l|c|c|c|c|c|c|c}\hline
~~~\textbf{Model}&~~~~~\textbf{B@1}~~~~&~~~~\textbf{B@2}~~~~&~~~~\textbf{B@3}~~~~&~~~~\textbf{B@4}~~~~
&~~~~\textbf{M}~~~~&~~~~\textbf{R}~~~~&~~~~\textbf{C}~~~~\\ \hline
~~~\textbf{{LSTM-T$_{frame}$}}                                & 12.71    & 7.24    & 4.01    & 1.99    & 8.99    & 14.67    & 13.82 \\
~~~\textbf{{LSTM-T$_{opt}$}}               & 12.46    & 7.08    & 3.96    & 1.97    & 8.72    & 14.55    & 13.60 \\\hline
~~~\textbf{{LSTM-T}} & 13.19    & 7.75    & 4.48    & 2.31    & 9.26    & 15.18    & 14.97 \\\hline

\end{tabular}
\end{table*}

In particular, given the input video with frame and optical flow sequences, each input frame/optical flow sequence ($\{{{\bf{v}}^{\left( 1 \right)}_i}\}^{K}_{i=1}$) is fed into a two-layer LSTM with attention mechanism. At each time step $t$, the attention LSTM decoder firstly collects the maximum contextual information by concatenating the input word $w_t$ with the previous output of the second-layer LSTM unit ${\bf{h}}^2_{t-1}$ and the mean-pooled video-level representation {\small $\overline {\bf{v}}  = \frac{1}{K}\sum\limits_{i = 1}^K {{\bf{v}}_i^{\left( 1 \right)}}$}, which will be set as the input of the first-layer LSTM unit. Hence the updating procedure for the first-layer LSTM unit is as
\begin{equation}\label{Eq:Eq5}\small
{\bf{h}}_t^1 = {f_1}\left( {\left[ {{\bf{h}}_{t - 1}^2,{{\bf{W}}_s}{{\bf{w}}_t},\overline {\bf{v}} } \right]} \right),
\end{equation}
where ${\bf{W}}_s\in {{\mathbb{R}}^{{D^1_s}\times {D_s}}}$ is the transformation matric for input word $w_t$, ${\bf{h}}_t^1 \in {{\mathbb{R}}^{D_h}}$ is the output of the first-layer LSTM unit, and $f_1$ is the updating function within the first-layer LSTM unit. Next, depending on the output ${\bf{h}}_t^1$ of the first-layer LSTM unit, a normalized attention distribution over all the frame/optical flow features is generated as:
\begin{equation}\label{Eq:Eq6}\small
\begin{array}{l}
a_{t,i}={\bf{W}}_a\left[\tanh\left({\bf{W}}_f{{\bf{v}}_i^{\left( 1 \right)}} + {\bf{W}}_h{\bf{h}}_t^1\right)\right],\\
\lambda_t=softmax\left({\bf{a}}_t\right),
\end{array}
\end{equation}
where $a_{t,i}$ is the $i$-th element of ${\bf{a}}_t$, ${\bf{W}}_a\in {{\mathbb{R}}^{{1}\times {D_a}}}$, ${\bf{W}}_f\in {{\mathbb{R}}^{{D_a}\times {D_v}}}$ and ${\bf{W}}_h\in {{\mathbb{R}}^{{D_a}\times {D_h}}}$ are transformation matrices. $\lambda_t \in\mathbb R^{K}$ denotes the normalized attention distribution and its $i$-th element $\lambda_{t,i}$ is the attention probability of ${\bf{v}}_i^{\left( 1 \right)}$. Based on the attention distribution, we calculate the attended video-level representation {\small $\hat {\bf{v}}_t  = \sum\limits_{i = 1}^K {\lambda_{t,i}{\bf{v}}_i^{\left( 1 \right)}}$} by aggregating all the frame/optical flow features weighted with attention.  We further concatenate the attended video-level feature $\hat {\bf{v}}_t$ with ${\bf{h}}_t^1$ and feed them into the second-layer LSTM unit, whose updating procedure is thus given as:
\begin{equation}\label{Eq:Eq7}\small
{\bf{h}}_t^2 = {f_2}\left( {\left[ {\hat {\bf{v}}_t,{\bf{h}}_{t}^1} \right]} \right),
\end{equation}
where $f_2$ is the updating function within the second-layer LSTM unit. The output of the second-layer LSTM unit ${\bf{h}}_t^2$ is leveraged to predict the next word $w_{t+1}$ through a softmax layer. Note that the policy gradient optimization method with reinforcement learning \cite{Liu:2016PGSPIDEr,Rennie:2016SCST} is additionally leveraged to boost the sentence generation performances specific to METEOR metric.

\section{Experiments}

\subsection{Features and Parameter Settings}
Each word in the sentence is represented as ``one-hot" vector (binary index vector in a vocabulary). For the input video representations, we take the output of 2048-way $pool5$ layer from P3D ResNet \cite{qiu2017learning} pre-trained on Kinetics dataset \cite{kay2017kinetics} as frame/optical flow representation. The dimension of the hidden layer in each LSTM $D_h$ is set as 1,000. The dimension of the hidden layer for measuring attention distribution $D_a$ is set as 512.

\subsection{Results}
Two slightly different settings of our LSTM-T are named as LSTM-T$_{frame}$ and LSTM-T$_{opt}$ which are trained with only frame and optical flow sequence, respectively. Table \ref{table:tabdense} shows the performances of our models on ActivityNet captions validation set. The results clearly indicate that by utilizing both frame and optical flow sequences in a late fusion manner, our LSTM-T boosts up the performances.

\section{Conclusions}
In this challenge, we mainly focus on the dense-captioning events in videos task and present a system by leveraging the three-stage workflow for temporal event proposal and LSTM-based captioning model with temporal attention mechanism for sentence generation. One possible future research direction would be how to end-to-end formulate the whole dense-captioning events in videos system.

{
\bibliographystyle{ieee}
\bibliography{egbib}
}

\end{document}